\pdfoutput=1

\documentclass[11pt]{article}

\usepackage[final]{acl}

\usepackage{times}
\usepackage{latexsym}

\usepackage[T1]{fontenc}

\usepackage[utf8]{inputenc}

\usepackage{microtype}

\usepackage{inconsolata}
\usepackage{amsthm, amsmath, amssymb}
\usepackage{bbm}
\usepackage{booktabs}
\usepackage{caption}
\usepackage{subcaption}
\usepackage{nameref}

\usepackage{algorithm}
\usepackage{algpseudocode}
\usepackage{caption}
\usepackage{graphicx}

\newtheorem{definition}{Definition}

\newtheorem*{theorem*}{Theorem}

\newcommand{\paren}[1]{\left( #1\right)}

\newcommand{\one}{\mathbbm{1}}

\newcommand{\yo}{y^{(0)}}

\newcommand{\yt}{y^{(1)}}

\newcommand{\zo}{z^{(0)}}
\newcommand{\zt}{z^{(1)}}

\newcommand{\br}[1]{\left[ #1\right]}

\title{Bayesian Calibration of Win Rate Estimation with LLM Evaluators}

\author{Yicheng Gao\thanks{Equal contribution}$^{1}$ \quad
  Gonghan Xu$^{*1}$\quad
  Zhe Wang$^{1}$ \quad
  Arman Cohan$^{1}$ \vspace{4pt}\\
  $^1$Yale University \vspace{4pt}\\
  \texttt{\{charlie.gao, gonghan.xu, zhe.wang.zw439, arman.cohan\}@yale.edu} \\}

\begin{document}
\maketitle

\begin{abstract}
Recent advances in large language models (LLMs) show the potential of using LLMs as evaluators  for assessing the quality of text generations from LLMs. However, applying LLM evaluators naively to compare or judge between different systems can lead to unreliable results due to the intrinsic win rate estimation bias of LLM evaluators. In order to mitigate this problem, we propose two calibration methods, Bayesian Win Rate Sampling (BWRS) and Bayesian Dawid-Skene, both of which leverage Bayesian inference to more accurately infer the true win rate of generative language models. We empirically validate our methods on six datasets covering story generation, summarization, and instruction following tasks. We show that both our methods are effective in improving the accuracy of win rate estimation using LLMs as evaluators, offering a promising direction for reliable automatic text quality evaluation.
\end{abstract}

\section{Introduction}

\begin{figure*}[t]
    \centering
    \includegraphics[width=\linewidth]{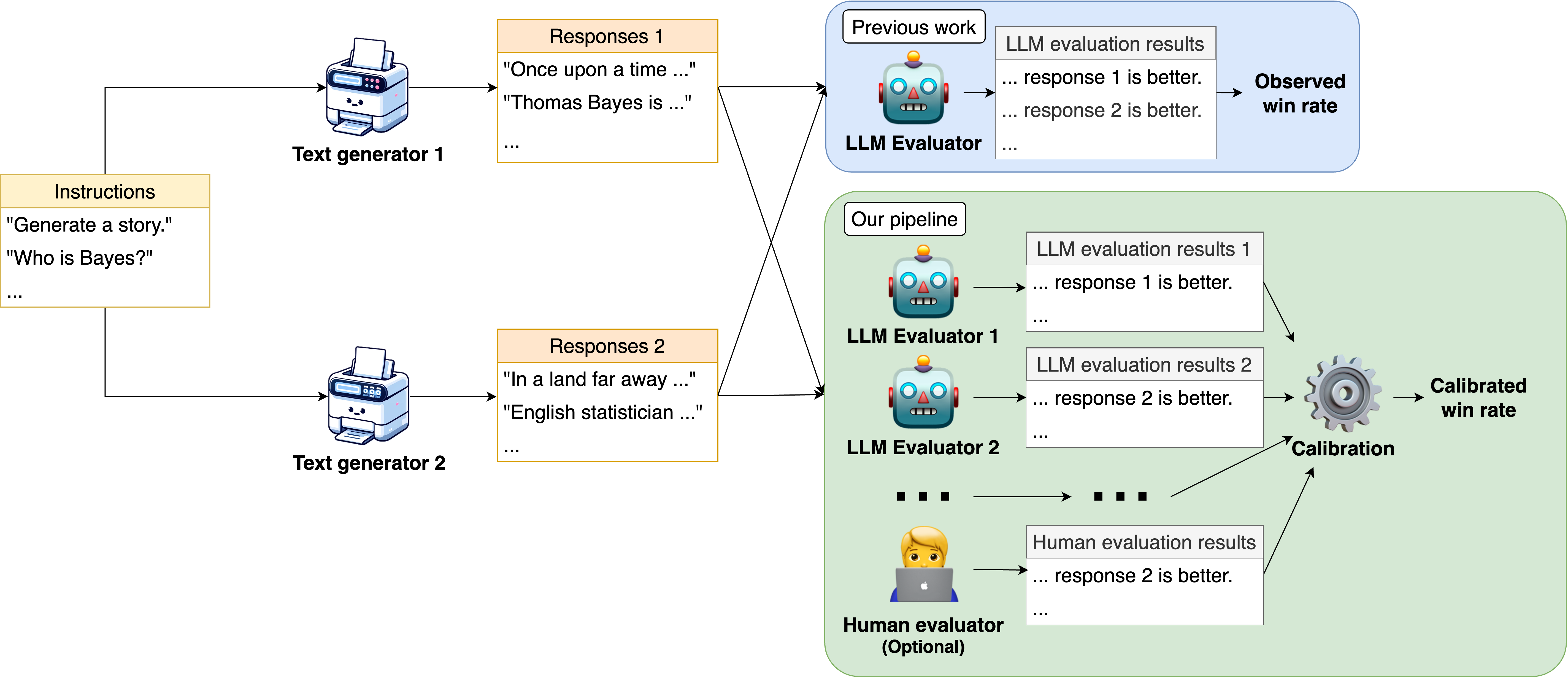}
    \caption{Illustration of our pipeline and previous work. The ``calibration'' part of our pipeline indicates one of BWRS or Bayesian Dawid-Skene.}
    \label{fig:pipeline}
\end{figure*}

Evaluating the quality of AI-generated text has been a longstanding and evolving challenge in NLP. In recent years, this challenge has become increasingly crucial due to the growing interest in the field of generative AI. While human judgment is still considered the most reliable form of assessment, common automatic approaches to evaluating quality of AI-generated text include heuristic-based evaluation metrics \cite{BLEU, rouge, mauve}, model-based evaluation metrics \cite{BERTScore, QAFactEval, alignscore, MENLI}, and recently, LLM-based evaluations \cite{prometheus, prometheus2, pandalm2024}. Due to their relative low cost and high correlation with human preferences, LLM-based evaluations (aka LLM-as-a-judge) are receiving increasing attention. Most previous studies that apply LLM evaluators \cite{chiang-lee, chiang-lee-2023-closer, alpaca, prometheus, prometheus2, pandalm2024, liu2024reife} attempt to improve the agreement between LLM evaluators and human preference by training expert models for evaluation or improving prompting strategies. However, such methods often either require compute-expensive finetuning, or suffer from common problems of LLM evaluators such as position bias \cite{llm-not-fair}, self-preference, and more \cite{llm-eval-bias}. Besides, as we will discuss in Section \ref{subsec:direct-k-as-p}, directly applying a non-perfect LLM evaluator will result in a bias problem in the estimation of win rate.

In this paper, we attempt to address these challenges by proposing two methods, Bayesian Win Rate Sampling (BWRS) and Bayesian Dawid-Skene. A general illustration of our pipeline is shown in Figure \ref{fig:pipeline}. Our approaches leverage Bayesian inference to enhance the accuracy of win rate estimations between competing text generators using evaluation results of LLM evaluators and sparse or no human evaluation data.
By employing these methods, we observe a closer alignment between LLM and human judgment in terms of win rate between two text generator models. Our results on six diverse datasets demonstrate that both BWRS and Bayesian Dawid-Skene effectively reduce win rate estimation bias of LLM evaluators, marking a promising step toward more trustworthy automatic evaluations in NLP. \footnote{The code and data used in our experiments are available at \url{https://github.com/yale-nlp/bay-calibration-llm-evaluators} under Apache 2.0 license.} %
The contribution of this paper is threefold:
\begin{itemize}
    \item We identify and formulate the win rate estimation bias problem associated with LLM evaluators.
    \item We conduct exploratory study on mitigating this bias with Bayesian inference. Specifically, we propose BWRS and Bayesian Dawid-Skene, both of which are shown effective in calibrating win rate estimation given LLM evaluation results, and optionally, some human evaluation results. 
    \item We publish our LLM evaluation annotations to facilitate future study in LLM-based evaluation.
\end{itemize}

\section{Related work}
\paragraph{LLM as evaluators}\label{subsec:llm-as-evaluators}

A line of research in LLM-based evaluation evaluated the performance of LLM evaluators and proposed methods to improve them. Some works applied various prompting techniques to improve the accuracy of LLM evaluation, including chain of thought \cite{liu-etal-2023-g}, evaluation with explanation \cite{chiang-lee-2023-closer}, multi-LLM discussion \cite{chan2023chateval, li2023prd}, calibration with human expert \cite{liu2023calibrating} and active optimization of evaluation protocol \cite{xu2024largelanguagemodelsactive}. Some other works \cite{pandalm2024, prometheus, prometheus2} trained expert models in evaluation. As for evaluating the general capability of LLM evaluators, most previous studies \cite{liu-etal-2023-g, chiang-lee, chiang-lee-2023-closer, alpaca, liu2024reife, liusie-etal-2024-llm, thakur2024judgingjudgesevaluatingalignment} used correlation coefficients such as Pearson's correlation and Kendall's tau or annotator agreement coefficients such as Cohen's kappa and Scott's pi to measure the preference of different LLM evaluators compared with human evaluators. 

On the application side, LLM evaluators are often applied to build LLM rankings. AlpacaFarm \cite{alpaca} proposed a simple LLM evaluation framework by looking at the win rate decided by a strong LLM evaluator (i.e., GPT-4) on a large number of texts generated by the two generators under the same generation prompts. Auto-Arena \cite{zhao2024auto} used LLM judge agents to determine the winner of each LLM pair. However, as we'll discuss in Section \ref{subsec:direct-k-as-p}, these methods can lead to biased win rate estimations, especially when the LLM evaluators do not align well enough with human preferences. 

\paragraph{Annotation models}\label{subsec:bay-annotation-models}

In the field of crowdsourced annotations, a line of research focuses on simultaneously modeling the accuracy of individual annotators and determining the true labels of tasks. These works mostly target aggregating crowdsourced data and improving data quality in case of non-expert or adversarial annotators. Dawid-Skene \cite{dawid1979maximum} was the first model proposed to consider individual annotator error rates by using maximum likelihood estimation to infer true labels from annotators with different accuracies. Since then, many other models \cite{albert2004cautionary, carpenter2008multilevel, GLAD, pmlr-v22-kim12, hovy-etal-2013-learning, passonneau-carpenter-2014-benefits, zhang2016spectral} were developed to improve performance and efficiency. These methods were originally proposed to model the accuracy of human annotators, in our paper we instead apply them to model LLM evaluators.

Some concurrent works also explored methods for aggregating annotations from multiple LLMs. \citet{10.1007/978-981-97-3076-6_12} used majority voting and the Dawid-Skene model to aggregate judgment results from multiple LLMs on the Legal Textual Entailment task. LLM-Ensemble \cite{10.1145/3626772.3661357} proposed an ensemble method based on the Dawid-Skene model for attribute value extraction with multiple LLMs. \citet{yao2024a} extended the Dawid-Skene model to estimate uncertainty and aggregate final answers from multiple LLMs in question-answering tasks. However, all these concurrent works focus on aggregating LLM evaluation results on specific tasks. To the best of our knowledge, our work is the first providing an analysis and solution on improving win rate estimation using LLM evaluators under a general text generation task scenario.

\section{Methods}

In this section, we first formalize the win rate estimation bias problem associated with directly applying LLM evaluator results, and then propose our methods to address this problem. In general, our methods attempt to derive more accurate estimators of the relative win rate between text generators by statistical calibration techniques and integrating optional human-based prior knowledge. Ultimately, we are able to improve win rate estimation accuracy without the need for costly, large-scale human annotations.

\subsection{Problem formalization}\label{subsec:problem formalization}
\subsubsection{True win rate and observed win rate}
Consider two LLMs as text generators (LLM generators) $G_0$ and $G_1$. Let $\Sigma$ be the set of all possible inputs to the text generators, and let $\Omega$ be the set of all possible outputs given the inputs from $\Sigma$. We can then define the LLMs as two functions $G_0 : \Sigma \rightarrow \Omega$ and $G_1 : \Sigma \rightarrow \Omega$. Additionally, let $P_\Sigma$ be a probability distribution on $\Sigma$ that denotes the probability of each input to appear, let $\sigma\sim P_\Sigma$ be a random input.

Let $H: \Omega \times \Omega \rightarrow \{0, 1\}$ be the \textbf{average human evaluator function}, which assesses the relative quality of two outputs. $H(y_0, y_1) = 0$ indicates that the output $y_0$ is preferred over $y_1$ by an average human expert (we assume that ``average human expert" exists), and $H(y_0, y_1) = 1$ indicates the opposite.
Let $T_e : \Omega \times \Omega \rightarrow \{0, 1\}$ be the \textbf{LLM evaluator function}, which represents the preference of a certain LLM evaluator $e$. Let $P$ be a probability measure that encapsulates the stochastic nature of $\sigma$, $G_0$, $G_1$, $H$, and $T_e$.

Given the notations above, we define the following variables:

\begin{definition}[True win rate]\label{def:p}
    The true win rate $p$ is defined as:
    \begin{equation}
    p \triangleq P\paren{H(G_0(\sigma), G_1(\sigma)) = 0}
    \end{equation}
\end{definition}

\begin{definition}[Observed win rate]\label{def:k}
    The observed win rate $k$ of an LLM evaluator $e$ is defined as:
    \begin{equation}
    k_e \triangleq  P\paren{T_e(G_0(\sigma), G_1(\sigma)) = 0}
    \end{equation}
\end{definition} 

Intuitively, the true win rate $p$ is the probability that $G_0$ will generate a ``truly better'' output than $G_1$ when they are given the same, arbitrary input, where ``truly better'' means being regarded as ``better'' by a human expert on average. Similarly, the observed win rate $k$ is the probability that $G_0$ will be evaluated by an LLM evaluator as generating a better output than $G_1$ when they are given the same, arbitrary input.

Due to the complexity of the stochasticity in $p$ and $k_e$, it is unrealistic to derive them analytically. However, given a large number of input-output pairs evaluated by human and LLM evaluators, we can approximate $p$ and $k_e$ empirically. We formalize it as follows.

Assume $n$ is a large number. Then for $n$ outputs $\yo_i \,(i \in [n])$ 
generated by 
$G_0$ and $n$ outputs $\yt_i \,(i \in [n])$ generated by $G_1$ given the same set of $n$ inputs of interest, we let a human evaluator $h$ and the LLM evaluator $e$ carry out $n$ comparison tasks, where the $i$-th comparison task is between $\yo_i$ and $\yt_i$. Then the true win rate $p$ and the observed win rate $k_e$ can be empirically approximated with
\begin{equation}
    \hat{p}=\frac{1}{n}\sum_{i=1}^n \br{1-H_h(\yo_i, \yt_i)}
    \label{eqn:emp_p}
\end{equation}

\begin{equation}
    \hat{k}_e=\frac{1}{n}\sum_{i=1}^n \br{1-T_e(\yo_i, \yt_i)}
    \label{eqn:emp_k}
\end{equation}
where $H_h: \Omega \times \Omega \rightarrow \{0, 1\}$ is the human evaluator function of a specific human evaluator $h$ (or an aggregation of multiple human evaluators). Note that in our experiments, in order to make sure that $\hat{p}$ is an accurate estimator of $p$, we assume that the preference of $h$ is representative of an average human expert evaluator.

\subsubsection{Evaluator accuracy}

We also define two variables $q_0^e$ (\textbf{true positive evaluation accuracy}) and $q_1^e$ (\textbf{true negative evaluation accuracy}) associated with an LLM evaluator $e$\footnote{For simplicity, we will use ``evaluator accuracies'' when we refer to $q_0^e$ and $q_1^e$ together. }. Given two arbitrary outputs generated under the same arbitrary input where the first output is evaluated as ``better'' than the second one by an average human expert, $q_0^e$ is defined as the conditional probability that $e$ will give the same evaluation as an average human expert. In other words, we have 
\begin{align}
    q_0^e\triangleq  P(&T_e(G_0(\sigma), G_1(\sigma))=0\mid\notag\\ &H(G_0(\sigma), G_1(\sigma))=0)
    \label{eqn:q0}
\end{align}
where the random element $\sigma\in\Sigma$ and probability measure $P$ follow the same notions as in the definitions of $p$ and $k$. Similarly, we have
\begin{align}
    q_1^e\triangleq  P(&T_e(G_0(\sigma), G_1(\sigma))=1\mid\notag\\ &H(G_0(\sigma), G_1(\sigma))=1)
    \label{eqn:q1}
\end{align}
Empirically, we can approximate $q_0^e$ and $q_1^e$ with

\begin{equation}
    \hat{q_0^e}=\frac{\sum\limits_{i=1}^n\one\br{T_e(\yo_i, \yt_i)=H_h(\yo_i, \yt_i)=0}}{\sum_{i=1}^n\one(H_h(\yo_i, \yt_i)=0)}
    \label{eqn:emp_q0}
\end{equation}
where $\one(\cdot)$ is the indicator function. Similarly, we have
\begin{equation}
    \hat{q_1^e}=\frac{\sum\limits_{i=1}^n\one\br{T_e(\yo_i, \yt_i)=H_h(\yo_i, \yt_i)=1}}{\sum_{i=1}^n\one(H_h(\yo_i, \yt_i)=1)}
    \label{eqn:emp_q1}
\end{equation}

\begin{algorithm*}
\begin{algorithmic}[1]
\small
\caption{Bayesian Win Rate Sampling (BWRS) algorithm}
\algnewcommand{\LineComment}[1]{\State \(\triangleright\) #1}
\label{algorithm:bwrs}
\State \textbf{Input:} Target dataset without human annotation: $D = \{(\yo_i, \yt_i), i \in [n]\}$; reference dataset along with human annotation: $F = \{(\zo_i, \zt_i), i \in [m]\}$; annotation by a set of LLM evaluators $E = \{e_1, e_2, \ldots, e_{|E|}\}$ on $D$: $D_{E} = \{T_{e}(\yo_i, \yt_i), i \in [n], e \in E\}$; annotation by LLM evaluators $E$ on $F$: $F_E = \{T_{e}(\zo_i, \zt_i), i \in [m], e \in E\}$; annotation by human evaluator $h$ on $F$: $F_h = \{H_h(\zo_i, \zt_i), i \in [m]\}$; number of samples drawn for each evaluator: $N$
\State \textbf{Output:} An estimation of the true win rate $p$
\LineComment{Number of data points on $F$ with the same human evaluation result (0 or 1)}
\State $n_0 = |\{(\zo_i, \zt_i)\in F :\, H_h(\zo_i, \zt_i) = 0\}|$
\State $n_1 = |\{(\zo_i, \zt_i)\in F :\, H_h(\zo_i, \zt_i) = 1\}|$
\LineComment{Number of correct judgments by each $e \in E$ on $F$}
\For{$e \in E$}
    \State $s_{0}^{e} = |\{(\zo_i, \zt_i)\in F :\, H_h(\zo_i, \zt_i) = T_{e}(\zo_i, \zt_i) = 0\}|$
    \State $s_{1}^{e} = |\{(\zo_i, \zt_i)\in F :\, H_h(\zo_i, \zt_i) = T_{e}(\zo_i, \zt_i) = 1\}|$
    \State $n_{k}^{e} = |D|$
    \State $s_{k}^{e} = |\{(\yo_i, \yt_i)\in D :\, T_{e}(\yo_i, \yt_i) = 0\}|$
\EndFor
\State sample list = $\emptyset$
\For{$i = 1,2,...,N$}
    \For{$e \in E$}
        \LineComment{Estimated evaluator accuracies for $e$}
        \State Draw $q_0^{e} \sim \text{Beta}(s_0^{e} + 1, n_0 - s_0^{e} + 1)$
        \State Draw $q_1^{e} \sim \text{Beta}(s_1^{e} + 1, n_1 - s_1^{e} + 1)$
        \LineComment{Observed win rate for $e$}
        \State Draw $k_{e} \sim \text{Beta}(s_k^{e} + 1, n_k^{e} - s_k^{e} + 1)$
        \State Derive sample $\hat{p}_{e} = \frac{k_{e} + q_1^{e} - 1}{q_0^{e} + q_1^{e} - 1}$, append to sample list
    \EndFor
\EndFor
\State \Return mean ($\hat{p}_{mean}$) or mode ($\hat{p}_{mode}$) of KDE(sample list)
\end{algorithmic}
\end{algorithm*}

\subsubsection{Win rate estimation}

As we discussed in Section \ref{subsec:llm-as-evaluators}, the true win rate $p$ can be used as a metric to compare various generative LLMs. Specifically, for two generative LLMs $G_0$ and $G_1$, $G_0$ outperforms $G_1$ when $p > 0.5$. Conversely, $G_1$ outperforms $G_0$ when $p < 0.5$. Furthermore, the absolute value of $p$ signifies the degree of superiority of one LLM to another. Given a list of LLMs $\Gamma = [G_a, G_b, ...]$ of interest and a certain baseline generative LLM $G$, we can use the $p$ values of $G$ with respect to each generator in $\Gamma$ to compare the LLMs in $\Gamma$ (1 vs. n comparison). Therefore, it is a meaningful question to derive an accurate estimation of $p$. This is the essential goal of this paper.

\subsection{Estimation by observed win rate}
\label{subsec:direct-k-as-p}
A simple approach employed by prior work \cite{alpaca} to approximate $p$ is to directly apply the observed win rate $k_e$. Here we show that this approach suffers from a win rate estimation bias problem when the evaluator accuracies are not high enough.

By the Law of Total Probability we have
\begin{align}
\label{eq:kt-p-q}
    k_e = & P\paren{T_e(G_0(\sigma), G_1(\sigma)) = 0} \notag\\
    = & P(H(G_0(\sigma), G_1(\sigma))=0)\cdot q_0^e + \notag\\
    & P(H(G_0(\sigma), G_1(\sigma))=1)\cdot (1 - q_1^e) \notag\\
    = & pq_0^e + (1 - p)(1 - q_1^e)
\end{align}

Therefore, using $k_e$ to approximate $p$ will result in the following win rate estimation error:
\begin{align}
    |k_e - p| = & |pq_0^e + (1 - p)(1 - q_1^e) - p| \notag \\
    = & |pq_0^e + pq_1^e - 2p - q_1^e + 1|
\end{align}

We can see that $k_e = p$ only under very special conditions such as $q_0^e = q_1^e = 1$, which is typically not the case for LLM evaluators. In order to fix this win rate estimation bias problem, we propose the following two methods to improve the accuracy in the estimation of $p$.

\subsection{Bayesian Win Rate Sampling}
\label{subsec:bwrs}

First, we propose a sampling-based algorithm, Bayesian Win Rate Sampling (BWRS), which is shown in Algorithm \ref{algorithm:bwrs}. The intuition of the BWRS algorithm is that, given an LLM evaluator $e$ and a dataset $D = \{(\yo_i, \yt_i), i\in [n]\}$ containing outputs generated by $G_0$ and $G_1$ with respect to the same set of inputs, we first apply the LLM evaluator $e$ to generate its annotations $\{T_e(\yo_i, \yt_i), i\in [n]\}$ on $D$ and then apply Equation \ref{eqn:emp_k} to approximate the observed win rate, $k_e$. Next, assume we have access to some human annotations, either on a small fraction of $D$ or on a similar reference dataset $F$, then we are able to approximate $q_0^e$ and $q_1^e$ using Equation \ref{eqn:emp_q0} and \ref{eqn:emp_q1}. Finally, we apply the following equation rearranged from Equation \ref{eq:kt-p-q}:
\begin{equation}
    p = \frac{k_e + q_1^e - 1}{q_0^e + q_1^e - 1}
    \label{eqn:p}
\end{equation}
given the assumption that $q_0^e + q_1^e \neq 1$. \footnote{In practice, though this assumption is satisfied under most cases, some values of evaluator accuracies might cause sampling failure. Please refer to \nameref{sec:limitations} for details.} We can use the approximated values of $k_e$, $q_0^e$, and $q_1^e$ to derive one sample of $p$, which characterizes the relative performance between $G_0$ and $G_1$.

Note that there are still two key differences between the intuition above and our actual implementation described in Algorithm~\ref{algorithm:bwrs}. First, in our implementation, instead of estimating $k_e, q_0^e, q_1^e$ directly using Equations \ref{eqn:emp_k}, \ref{eqn:emp_q0}, \ref{eqn:emp_q1}, we use Bayesian inference and apply the Beta-Bernoulli model to estimate the posterior distributions for $k_e$, $q_0^e$, and $q_1^e$. Second, instead of using one evaluator model $e$, we use a set of LLM evaluators $E = \{e_1,e_2,...,e_{|E|}\}$. We aggregate the results by aggregating all the samples obtained using each LLM evaluator. Concretely, we obtain $N$ (10000 in our case) samples of $p$ for each LLM evaluator by sampling from the posterior distributions of $k_e$, $q_0^e$, $q_1^e$ and applying Equation \ref{eqn:p}. Then, we apply Kernel Density Estimation (KDE) on all the $p$ samples to approximate the distribution of $p$. Finally, we estimate the value of $p$ using the mean $\hat{p}_{mean}$ or mode $\hat{p}_{mode}$ of this distribution.
The purpose of applying a Bayesian setting is to incorporate the uncertainty of $k_e, q_0^e, q_1^e$ into consideration, and also facilitate the usage of prior knowledge on evaluator accuracies, which will be discussed in Section \ref{subsec:win-rate-experiments}. The purpose of using multiple LLM evaluators is to mitigate the effect of potential inaccurate estimation of individual LLM evaluators' evaluation accuracy.

\setcounter{algorithm}{0}
\makeatletter
\renewcommand{\ALG@name}{Model}
\makeatother
\algnewcommand{\LineComment}[1]{\State \(\triangleright\) #1}
\begin{algorithm}[ht]
\begin{algorithmic}[1]
\small
\caption{Bayesian Dawid-Skene model}
\label{algorithm:bayesian-ds-two-class}
    \LineComment{Prior class prevalence}
    \State Draw $p \sim \text{Beta}(\alpha_p, \beta_p)$ 
    \For{$e \in E$}
        \LineComment{Evaluator accuracies}
        \State Draw $q_0^e \sim \text{Beta}(\alpha_{q_0}, \beta_{q_0})$ 
        \State Draw $q_1^e \sim \text{Beta}(\alpha_{q_1}, \beta_{q_1})$ 
    \EndFor
    \For {$i = 1$ \textbf{to} $n$}
        \LineComment{Ground truth labels}
        \State Draw $h_i \sim \text{Bernoulli}(p)$ 
        \For{$e \in E$}
            \LineComment{Predicted labels}
            \If{$h_i = 1$} 
                \State Draw $t^e_i \sim \text{Bernoulli}(q_1^e)$ 
            \Else
                \State Draw $t^e_i \sim \text{Bernoulli}(1 - q_0^e)$ 
            \EndIf
        \EndFor
    \EndFor
\end{algorithmic}
\end{algorithm}

\subsection{Bayesian Dawid-Skene model}
\label{subsec:bay-ds}

The vanilla Dawid-Skene model \cite{dawid1979maximum} is optimized with the Expectation-Maximization (EM) algorithm. Following \citet{bay-ds}, we instead use a Bayesian Dawid-Skene model. The pseudocode of our model is shown in Model~\ref{algorithm:bayesian-ds-two-class}. The parameters in this model include $\alpha_p, \beta_p, \alpha_{q_0}, \beta_{q_0}, \alpha_{q_1}, \text{and} \beta_{q_1}$. We initialize the distribution of $p$ with a uniform distribution, and thus $\alpha_p, \beta_p$ are initialized as 1. The initialization of the other parameters will be discussed in Section~\ref{subsec:win-rate-experiments}. We apply the evaluation results of each LLM evaluator $e$ as observations $t^e_i$, and use Hamiltonian Monte Carlo (HMC) sampling to fit the model and sample from the posterior distribution of $p$. Similar to BWRS, we use the posterior mean ($\hat{p}_{mean}$) and posterior mode ($\hat{p}_{mode}$) as two estimators of $p$. In order to improve sampling efficiency, we employ NUTS sampler \cite{hoffman2011nouturn} and the Binary Gibbs-Metropolis sampler implemented in PyMC \cite{PyMC}. We tune and sample from the model with 4 chains, with 10000 tuning steps and 10000 sampling steps on each chain. On an AMD EPYC 7763 processor, comparing each generator pair takes around 10 minutes.

\section{Experiment Settings}
\subsection{Datasets}
\label{subsec:datasets}
The datasets we use in the experiments are HANNA \cite{chhun-etal-2022-human}, OpenMEVA-MANS \cite{guan-etal-2021-openmeva}, SummEval \cite{summeval}, LLMBar \cite{llmbar}, MT-Bench \cite{mt-bench}, and LLMEval$^2$ \cite{llmeval2}, covering tasks of story generation (HANNA, OpenMEVA-MANS), summarization (SummEval), and instruction following (the other three). All of them provide machine-generated content with human annotations. For MT-Bench and LLMEval$^2$, we used the smaller, curated versions prepared by the authors of the LLMBar paper \cite{llmbar}. For the three instruction following datasets, since they are presented as a list of (input, output1, output2, human preference) tuples without specifying which LLM generated each of the outputs, we simulate two LLM generators based on these datasets by randomly attributing 80\% of the human-preferred outputs to the (simulative) generator A and the rest 20\% to the (simulative) generator B such that the true win rate between them is 80\%. We chose the 80\%-20\% ratio to represent a substantial yet realistic performance difference between two models.

A detailed description about each dataset can be found in Appendix \ref{app:datasets}.

\subsection{Evaluator settings}
For HANNA, OpenMEVA-MANS, and SummEval, we prompt a set of LLM evaluators to compare the outputs of generator models in the datasets. Specifically, we employ GPT-3.5-turbo-0125 \cite{chatgpt} and Gemini-1.0-Pro \cite{Gemini} as the evaluator models for our experiments. GPT-3.5 has been proved to have positive correlation with human annotations \cite{chiang-lee, wang-2023-chatgpt}, while Gemini-1.0-Pro's performance on LLM evaluation have not yet been widely studied in previous works. For each output pair, we prompted each LLM evaluator to rate the two outputs that are based on the same input and generated by two different generator models. For each LLM evaluator, we used three prompting strategies including Score-only, Rate-explain, and Analyze-rate following \citet{chiang-lee-2023-closer}. For LLMBar, MT-Bench, LLMEval$^2$, the LLM evaluation work has already been carried out by \citet{llmbar}. For these three datasets, we selected the best LLM evaluators (GPT-4, PaLM 2, etc.) from the many ones used. More details regarding the specific LLM evaluator modes used for these datasets can be found in Appendix \ref{app:eval_acc_results}.

\subsection{Win rate estimation}
\label{subsec:win-rate-experiments}
After obtaining the human evaluation and LLM evaluation data, we apply BWRS (Section~\ref{subsec:bwrs}) and Bayesian Dawid-Skene (Section~\ref{subsec:bay-ds}) to each dataset described above. We conduct ``1 vs. n'' experiment on each dataset, where we select a baseline model (GPT-2) and compare its outputs to all the other text generators in the dataset. We employ this ``1 vs. n'' comparison strategy because the corresponding ``n vs. n'' strategy is much more costly in terms of computation time and budget. Additionally, we calculate the observed win rate $k$ by using Equation \ref{eqn:emp_k} and averaging over the results of all LLM evaluators combined. The error of estimating $p$ with the observed win rate (i.e., $|k - p|$) acts as a baseline that shows the aggregated performance of all the LLM evaluators applied without any calibration.

In order to further study the effectiveness of each estimation method, we also explore their performance given the following three different sources of human evaluation results. For simplicity, we refer to these human evaluation results as \textbf{prior}s, since they act as prior knowledge of human preferences in our methods.

\textbf{No prior}\footnote{The no prior setting is not applicable for BWRS, since BWRS requires informative priors of evaluator accuracies to be accurate.}\textbf{.} We assume no prior knowledge of evaluator accuracies, and only depend on the Dawid-Skene model to estimate the accuracy of each evaluator. In this case, we initialize the parameters of evaluator accuracies in Model~\ref{algorithm:bayesian-ds-two-class} with $\alpha_{q_0} = \alpha_{q_1} = 2, \beta_{q_0} = \beta_{q_1} = 1$, which is a beta distribution skewed towards higher $q_0$ and $q_1$ values, because we expect our evaluators to generally perform better than random guessing such that $q_0>0.5$ and $q_1>0.5$.

\textbf{In-distribution prior.} We assume that we have access to human evaluations on a subset of all output pairs generated by the two generators of interest. Then in Algorithm \ref{algorithm:bwrs} for BWRS, the reference dataset $F$ becomes a subset of $D$ and the human evaluation results are used as $F_h$ to obtain an estimate of each LLM evaluator's accuracies $q_0$, $q_1$. In the Bayesian Dawid-Skene model, the human evaluation results are instead used as observations ($h_i$ in Model \ref{algorithm:bayesian-ds-two-class}), while $\alpha_{q_0}$, $\beta_{q_0}$, $\alpha_{q_1}$, and $\beta_{q_1}$ are initialized in the same way as in the no prior setting. We refer to the ratio of human-evaluated output pairs over the entire dataset as \textbf{prior data ratio}. In our experiments, we try 10 different values of prior data ratio (0.1, 0.2, ..., 1.0) and compare the results.

\textbf{Out-of-distribution (OOD) prior.} We assume that we have access to human evaluations on some other reference dataset beyond the outputs generated by the two generators of interest. These human evaluation results are also used to calculate priors for $q_0$ and $q_1$. In our experiments, we use the generator pair in the reference dataset that has the closest observed win rate with the compared generators.
For BWRS, these priors are used as $F_e$ and $F_h$ in Algorithm~\ref{algorithm:bwrs}. For the Bayesian Dawid-Skene model, with the in-distribution prior setting, recall that the human evaluation priors are used as observations of ground truth labels $h_i$ in Model~\ref{algorithm:bayesian-ds-two-class}. For the OOD prior setting, they are instead only used to derive a prior distribution of the evaluator accuracies so that the model won't be affected as much by the distribution shift of evaluator accuracies on different generator models. Specifically, we use a Beta-Bernoulli model similar to the ones we used in BWRS. The only difference is that we normalize the Beta distribution parameters such that their average is 1 in order to prevent over-confident priors. Concretely, we initialize the distributions of $q_0^e$ and $q_1^e$ in Model~\ref{algorithm:bayesian-ds-two-class} for each evaluator $e$ as follows:

\begin{alignat}{2}
    &n_0 = &&| \{(\zo_i, \zt_i)\in \text{OOD} :\, H_h(\zo_i, \zt_i) = 0\} | \notag\\
    &n_1 = &&| \{(\zo_i, \zt_i)\in \text{OOD} :\, H_h(\zo_i, \zt_i) = 1\} | \notag\\
    &s_0 = &&| \{(\zo_i, \zt_i)\in \text{OOD} :\, \notag\\
    & && H_h(\zo_i, \zt_i) = T_e(\zo_i, \zt_i) = 0\} | \notag\\
    &s_1 = &&| \{(\zo_i, \zt_i)\in \text{OOD} :\, \notag\\
    & && H_h(\zo_i, \zt_i) = T_e(\zo_i, \zt_i) = 1\} | \notag\\
    \label{eq:bayds_ood_q_0}
    &q_0^e \sim && \text{Beta}(\frac{2s_0 + 2}{n_0 + 2}, \frac{2n_0 - 2s_0 + 2}{n_0 + 2}) \\
    \label{eq:bayds_ood_q_1}
    &q_1^e \sim && \text{Beta}(\frac{2s_1 + 2}{n_1 + 2}, \frac{2n_1 - 2s_1 + 2}{n_1 + 2}) 
\end{alignat}
where $\text{OOD}$ is the OOD set (dataset $F$) we use, the term $n_0 + 2$ and $n_1 + 2$ on the denominator of Equation~\ref{eq:bayds_ood_q_0} and~\ref{eq:bayds_ood_q_1} are both normalization terms as described above. We repeat the experiment ten times on each dataset for each prior setting.

\section{Results}

\begin{table*}[ht]
    \small
    \centering
    \begin{tabular}{c c c c c c}
    \toprule
    Evaluator model & Prompt template & $q_0\uparrow$ & $q_1\uparrow$ & $|q_0 - q_1|$ & Overall Accuracy $\uparrow$ \\
    \midrule
    Gemini-1.0-Pro & Score-only & 0.782 & 0.526 & 0.256 & 0.649 \\
    & Analyze-rate & \textbf{0.802} & 0.428 & 0.374 & 0.607 \\
    & Rate-explain & 0.760 & 0.512 & 0.248 & 0.631 \\
    GPT-3.5 & Score-only & 0.700 & 0.653 & 0.047 & \textbf{0.676} \\
    & Analyze-rate & 0.657 & \textbf{0.677} & 0.020 & 0.667 \\
    & Rate-explain & 0.699 & 0.655 & 0.044 & \textbf{0.676} \\
    \bottomrule
    \end{tabular}
    \caption{LLM evaluator accuracy with respect to human preferences when human preference is $0$ ($q_0$) or $1$ ($q_1$) across all pair-wise generator comparisons on HANNA, OpenMEVA-MANS, and SummEval. Best performance on each column is marked with bold font.}
    \label{tab:eval-accuracy}
\end{table*}

\begin{table*}[htbp]
    \small
    \centering
    \begin{tabular}{c c c c c}
    \toprule
    Dataset & Method & Prior setting & $|\hat{p}_{mean} - p|\downarrow$ & $|\hat{p}_{mode} - p|\downarrow$\\
    \midrule
    HANNA & Observed win rate (baseline) & / & \textbf{0.079} & \textbf{0.079} \\
    & Bayesian Dawid-Skene & No prior & 0.129 & 0.132 \\
    & Bayesian Dawid-Skene & OOD prior & 0.084 & 0.081 \\
    & BWRS & OOD prior & 0.129 & 0.095 \\
    \midrule
    OpenMEVA-MANS & Observed win rate (baseline) & / & 0.065 & 0.065\\
    & Bayesian Dawid-Skene & No prior & 0.065 & 0.065 \\
    & Bayesian Dawid-Skene & OOD prior & 0.034 & \textbf{0.033} \\
    & BWRS & OOD prior & 0.064 & 0.102 \\
    \midrule
    SummEval & Observed win rate (baseline) & / & 0.167 & 0.167 \\
    & Bayesian Dawid-Skene & No prior & 0.125 & 0.123 \\
    & Bayesian Dawid-Skene & OOD prior & 0.115 & \textbf{0.110} \\
    & BWRS & OOD prior & 0.112 & 0.112 \\
    \bottomrule
    \end{tabular}
    \caption{Results of win rate estimation with no prior and OOD prior on HANNA, OpenMEVA-MANS, and SummEval. Lower estimation bias ($|\hat{p}_{mean} - p|$ or $|\hat{p}_{mode} - p|$) is better. All results are averaged over all compared generator pairs in ten repetitive runs. The best estimator for each dataset is marked with bold font.}
    \label{tab:ood-no-prior-results}
\end{table*}

\begin{table*}[htbp]
    \small
    \centering
    \begin{tabular}{c c c c c}
    \toprule
    Dataset & Method & Prior setting & $|\hat{p}_{mean} - p|\downarrow$ & $|\hat{p}_{mode} - p|\downarrow$ \\
    \midrule
    LLMBar & Observed win rate (baseline) & / & 0.142 & 0.142 \\
    & Bayesian Dawid-Skene & No prior & 0.140 & \textbf{0.138} \\
    \midrule
    LLMEval$^2$ & Observed win rate (baseline) & / & 0.178 & 0.178 \\
    & Bayesian Dawid-Skene & No prior & 0.157 & \textbf{0.156} \\
    \midrule
    MT-Bench & Observed win rate (baseline) & / & \textbf{0.162} & \textbf{0.162} \\
    & Bayesian Dawid-Skene & No prior & 0.190 & 0.188 \\
    \bottomrule
    \end{tabular}
    \caption{Results of win rate estimation with no prior on the three instruction following datasets. Lower estimation bias ($|\hat{p}_{mean} - p|$ or $|\hat{p}_{mode} - p|$) is better. All results are averaged over all compared generator pairs in ten repetitive runs. The best estimator for each dataset is marked with bold font.}
    \label{tab:inst-following-no-prior-results}
\end{table*}

\begin{figure*}[ht]
    \centering
    \begin{subfigure}[b]{0.5\textwidth}
        \centering
        \includegraphics[width=\textwidth]{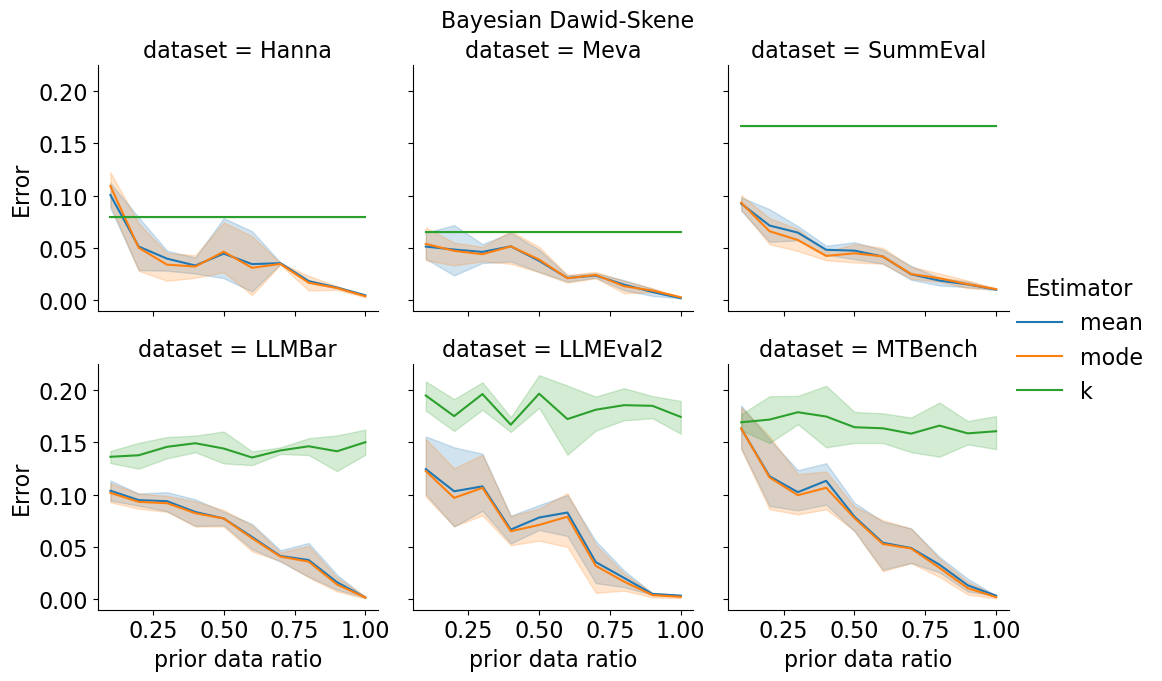} 
        \caption{Bayesian Dawid-Skene}
        \label{fig:bayds-in-dist}
    \end{subfigure}\hfill
    \begin{subfigure}[b]{0.5\textwidth}
        \centering
        \includegraphics[width=\textwidth]{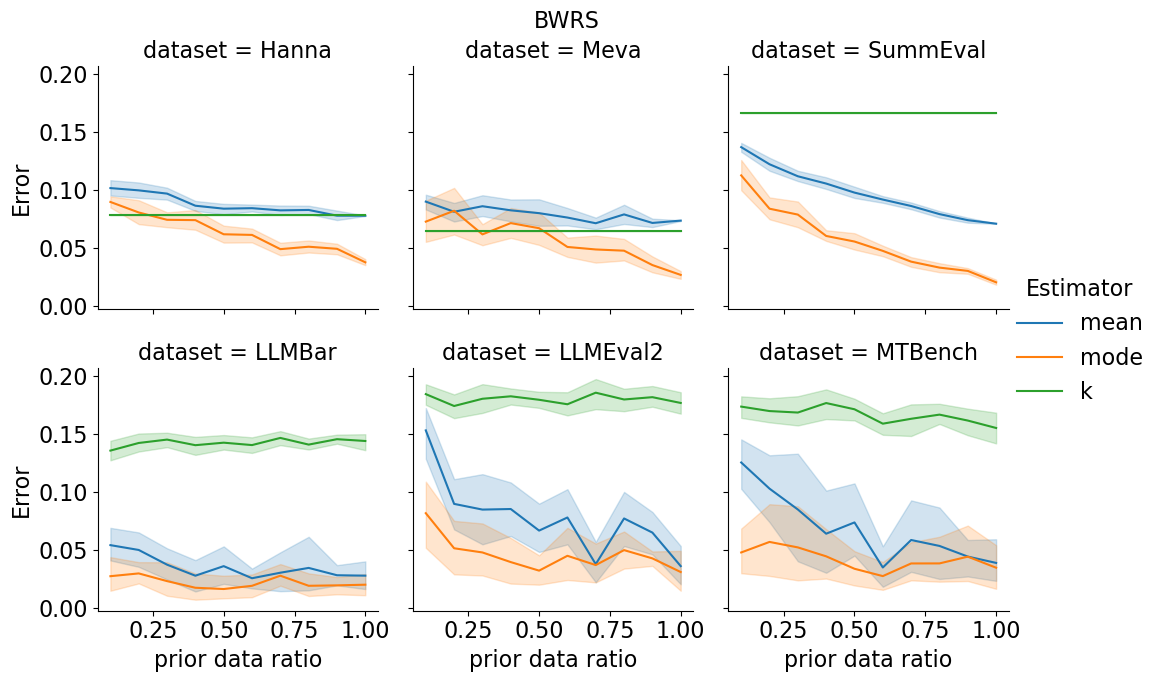}
        \caption{BWRS}
        \label{fig:bwrs-in-dist}
    \end{subfigure}
    \caption{Win rate estimation error with various proportions of the original data used as in-distribution prior. The results are averaged over all compared generator pairs. The mean and variance of all results are calculated over ten repetitive runs. The variance of $k$ values in the three instruction following datasets results from randomly assigning outputs to two simulative generators, as described in Section \ref{subsec:datasets}}
    \label{fig:in-dist}
\end{figure*}

In this section, we first analyze the evaluator accuracies on our datasets, and then list the results of our experiments, including win rate estimation with no prior, OOD prior, and in-distribution prior. We show that both our methods are able to effectively calibrate the estimation of win rate given good estimations of evaluator accuracies. We also show that even with no or OOD knowledge of human preference, our methods are still able to perform well overall.
\subsection{Evaluator accuracies}

\label{subsec:results-eval-acc}

For the three non-instruction following datasets (HANNA, OpenMEVA-MANS, SummEval) on which we carry out LLM evaluation by ourselves, the average accuracies of the LLM evaluators are shown in Table \ref{tab:eval-accuracy}. The overall accuracy is defined as the proportion of all pair-wise output comparisons where the LLM evaluation aligns with human evaluation. We can see that: 
\begin{itemize}
    \item In terms of overall accuracy, there is not a significant difference (>5\%) between the three prompt templates.
    \item There is a significant difference between $q_0$ and $q_1$ even though we applied the swap-and-sum strategy (see Appendix \ref{app:datasets}). This can be attributed to the correlation between evaluator accuracy and the difference between the generators' capabilities. When one generator is significantly better than the other, it is easier for the LLM evaluator to identify cases where the better generator does better, and harder when the better generator does worse. Also, Gemini-1.0-Pro evaluators suffer from this problem more significantly than GPT-3.5 evaluators. This shows the necessity of modeling $q_0$ and $q_1$ separately for each evaluator when comparing two generators.
\end{itemize}

For the instruction following datasets (LLMBar, LLMEval$^2$, MT-Bench), the overall evaluator accuracies are given in the LLMBar paper \cite{llmbar}, where the overall evaluation accuracies are generally above 70\% for the evaluator modes we use.

\subsection{Win rate estimation results}

The results of win rate estimation with no prior and OOD prior on HANNA, OpenMEVA-MANS, and SummEval are shown in Table \ref{tab:ood-no-prior-results}. We can observe that:
\begin{itemize}
    \item The mode estimator in Bayesian Dawid-Skene with OOD prior is the overall best estimator. In this setting, estimation of $p$ is more accurate than baseline ($k$) in all datasets except HANNA.
    \item The Bayesian Dawid-Skene model with OOD prior is more accurate than the model with no prior. This shows that the OOD prior is able to provide some useful information on the accuracy of each evaluator, which helps the Bayesian model converge to a better result.
\end{itemize}

The results of win rate estimation with no prior on LLMBar, LLMEval$^2$, and MT-Bench are shown in Table \ref{tab:inst-following-no-prior-results}. Note that OOD prior is not applicable for these instruction following datasets due to the absence of relevant data to act as the OOD set. We can see that the mode estimator in Bayesian Dawid-Skene with no prior outperforms the baseline in all datasets except MT-Bench.

The results of BWRS and Bayesian Dawid-Skene with in-distribution prior are shown in Figure \ref{fig:in-dist}. We can observe the following:
\begin{itemize}
    \item As prior data ratio increases, win rate estimation accuracy of both BWRS and Bayesian Dawid-Skene improves. This enhancement arises because having more human annotations for in-distribution data allows for a more precise assessment of evaluator accuracies and consequently leads to a more accurate estimation of the true win rate $p$. This shows that our methods will indeed offer a more accurate estimation of the true win rate $p$ if we have good estimations of $q_0$ and $q_1$. 
    \item The mode estimator shows consistently better performance compared with the mean estimator and $k$. 
    \item The proportion of human evaluation data needed to ensure improvement of the true win rate estimation varies for each dataset due to the internal variance of evaluator accuracies. Generally, a prior data ratio of 30\% would be sufficient for both Bayesian Dawid-Skene and BWRS, with one exception (BWRS for OpenMEVA-MANS).
\end{itemize}

\section{Conclusion}
In this paper, we identified and formulated the win rate estimation bias problem in using LLMs as evaluators to compare text generators, where discrepancies between non-perfect LLM evaluators and human preferences could lead to errors in win rate estimation. We proposed two methods, Bayesian Win Rate Sampling (BWRS) and Bayesian Dawid-Skene, in order to address this issue. We then obtained LLM evaluation results on six diverse datasets, and used these results to examine the effectiveness of our methods empirically. Our results showed that both BWRS and Bayesian Dawid-Skene can effectively mitigate the LLM evaluators' win rate estimation bias, especially given good approximations on evaluator accuracies. Our results also showed that even without in-distribution prior knowledge of human preferences, our methods are still able to effectively calibrate win rate estimation under most cases. The effectiveness of our methods manifests the possibility to calibrate win rate estimation in a post-hoc manner after LLM evaluations are completed, and also enlightens future study on applying annotation models for accurate win rate estimation using LLM evaluators.

\section*{Limitations}
\label{sec:limitations}

There are some limitations of our work. First, due to budget limit, for the non-instruction following datasets, we only examined our methods with GPT-3.5 and Gemini-1.0-Pro as LLM evaluators. Although we did incorporate more advanced LLM evaluators such as GPT-4 and PaLM 2 on the instruction following datasets, it would be illuminating to examine how more advanced evaluator models would affect our methods' performance on the non-instruction following datasets.

Second, the performance of both methods with OOD prior largely depends on the quality of OOD data. Specifically, when there is a large difference between evaluator accuracies on the OOD set and on the original dataset, our methods may produce highly-biased results. Therefore, in cases where human evaluation results on datasets with similar observed win-rates are absent, we would recommend against using OOD prior.

This paper is an exploratory study on adjusting the win rate estimation bias of LLM evaluators. Besides resolving the limitations above, the exploration in this field could also be extended in the following aspects:
\begin{itemize}
    \item Applying more complex annotator models. As discussed in Section~\ref{subsec:bay-annotation-models}, the Dawid-Skene model is the earliest annotator model proposed, and several improvements have been proposed since then. These improved methods can potentially lead to more accurate win rate estimation.
    \item Introducing more robust methods. The performance of our proposed methods is contingent upon the accuracy of LLM evaluators. Concretely, from Equation \ref{eqn:p} we know that

    \begin{align}\label{eq:p-legal-cond}
    \small
        0 < p < 1 \Leftrightarrow 
        \begin{cases}
            1 - q_1^e < k_e < q_0^e, & q_0^e+q_1^e > 1 \\
            q_0^e < k_e < 1 - q_1^e, & q_0^e+q_1^e < 1
        \end{cases}
    \end{align}

    We can see that, in order to make sure $p\in[0,1]$, the evaluator accuracies $q_0^e$ and $q_1^e$ must satisfy one of the conditions in Equation \ref{eq:p-legal-cond}. In cases where neither condition is satisfied, our methods can become unstable, and is prone to produce $p$ distributions with high bias and/or variance. We leave it for future research to propose methods that work well for LLM evaluators with low or unstable accuracies.
\end{itemize}

\bibliography{custom}

\appendix

\appendix

\section{Dataset details}
\label{app:datasets}
\textbf{HANNA} \cite{chhun-etal-2022-human} includes 1056 stories annotated by human raters with a 5-point Likert scale on 6 criteria: Relevance, Coherence, Empathy, Surprise, Engagement, and Complexity. These 1056 stories are based on 96 story prompts from the WritingPrompts \cite{fan-etal-2018-hierarchical} dataset. For each story prompt, HANNA collects 11 stories generated by 10 different generation models and a human, respectively. For our purpose of comparing automatic text generation systems, we did not use the stories written by humans in our experiments.
    
\textbf{OpenMEVA-MANS} \cite{guan-etal-2021-openmeva} is a sub-dataset within the OpenMEVA dataset. It contains 1000 stories generated by 5 generation models based on 200 prompts from WritingPrompts \cite{fan-etal-2018-hierarchical}. The overall quality of each story is rated by five humans on a 5-point Likert scale.
    
\textbf{SummEval} \cite{summeval} includes 1600 summaries annotated by human expert annotators with a 5-point Likert scale on 4 criteria: coherence, consistency, fluency, and relevance. These 1600 summaries are based on 100 source articles from the CNN/DailyMail dataset \cite{cnn}. For each source article, SummEval collects 16 summaries generated respectively by 16 different automatic summary generation systems. Each summary is scored by three human expert annotators.

\textbf{LLMBar} \cite{llmbar} consists of 419 instances, each containing an instruction paired with two outputs: one that faithfully follows the instruction and another that deviates from it but may possess superficially appealing qualities. The dataset is divided into two main parts: the Natural set, which includes instances from existing human-preference datasets that have been filtered and modified to ensure objective preferences, and the Adversarial set, which contains outputs crafted to mislead evaluators by emphasizing superficial qualities. LLMBar aims to provide a more rigorous and objective evaluation of LLM evaluators compared to previous benchmarks, achieving a high inter-annotator agreement rate of 94\% \cite{llmbar}.

\textbf{MT-Bench} \cite{mt-bench} comprises 80 questions as well as answers to these questions generated by six models. For each question and each pair of models, an evaluation task was constructed, totaling 1200 tasks. The actual dataset that we used is a subset of the original MT-Bench dataset curated by \citet{llmbar}, to construct which they labelled a human-preferred answer for each task using majority vote, removed all the ``tie'' instances, and then randomly sampled 200 instances. We found that five instances of this curated subset repeated themselves once, so we further removed the repeated ones and used the remaining 195 instances for our experiments.

\textbf{LLMEval$^2$} \cite{llmeval2}, similar to MT-Bench, is a question answering dataset where each instance comprises a question and two answers to that question. It consists of 2553 instances, each annotated with human preferences. The actual dataset that we used is a subset of the original LLMEval$^2$ dataset \cite{llmeval2} curated by \citet{llmbar}, to construct which they removed all the ``tie'' instances and then randomly sampled 200 instances.

For each dataset with multiple human evaluations on each piece of generated text, we averaged the human evaluation scores as the final human evaluation score for each piece of text.

\section{Evaluator setup details}
\label{app:eval_acc_results}

We prepared prompt templates into which the input and the two outputs would be inserted. 
Specifically, we used the following three prompting strategies following \citet{chiang-lee-2023-closer}.

The \textbf{Score-only} prompting strategy asks the LLM evaluator to only output the attribute scores of the generated texts without any further explanations.

The \textbf{Rate-explain} prompting strategy asks the LLM evaluator to rate the generated texts first and then provide an explanation for its ratings.

The \textbf{Analyze-rate} prompting strategy asks the LLM evaluator to first analyze the generated texts and then give the ratings for them.

Additionally, it has been reported that LLM evaluators suffer from position bias \cite{llm-not-fair}, meaning that their decisions are often falsely correlated with the order of presenting the compared texts. In order to address this problem, we employ a straightforward \textbf{swap-and-sum} strategy inspired by the LLMBar paper \cite{llmbar}. For each pair of outputs to be compared, we query the LLM evaluator twice with the original and swapped ordering of the outputs. We then sum the scores given by the LLM evaluator in the two queries and choose the generated text with the higher total score as the LLM-evaluated winner. In cases where the total score is even for both outputs, we consider their quality to be equal, and randomly select one as the winner.

The details of the LLM evaluator modes used by our experiments can be found in Tables \ref{tab:full-eval-accuracy} and \ref{tab:llm-eval-accuracy}. For the prompting templates used for the three instruction following datasets shown in Table \ref{tab:llm-eval-accuracy}, please refer to the LLMBar paper \cite{llmbar} for detailed explanations.

\begin{table*}[ht]
    \small
    \centering
    \begin{tabular}{c c c}
    \toprule
    Dataset & Evaluator model & Prompt template\\
    \midrule
    HANNA & GPT-3.5 Turbo & Score-only \\
          & & Rate-explain \\
          & & Analyze-rate \\
          & Gemini-1.0-Pro & Score-only \\
          & & Rate-explain \\
          & & Analyze-rate \\
    \midrule
    OpenMEVA-MANS & GPT-3.5 Turbo & Score-only \\
         & & Rate-explain \\
         & & Analyze-rate \\
         & Gemini-1.0-Pro & Score-only \\
         & & Rate-explain \\
         & & Analyze-rate \\
    \midrule
    SummEval & GPT-3.5 Turbo & Score-only \\
             & & Rate-explain \\
             & & Analyze-rate \\
             & Gemini-1.0-Pro & Score-only \\
             & & Rate-explain \\
             & & Analyze-rate \\
    \bottomrule
    \end{tabular}
    \caption{LLM evaluator modes used for the story generation and summarization datasets in our experiments.}
    \label{tab:full-eval-accuracy}
\end{table*}

\begin{table*}[ht]
    \small
    \centering
    \begin{tabular}{c c c}
    \toprule
    Dataset & Evaluator model & Prompt template \\
    \midrule
    LLMBar & GPT-4 & CoT \\
           & & Metrics \\
           & & Metrics Reference \\
           & & Reference \\
           & & Swap \\
           & & Swap CoT \\
           & & Vanilla \\
           & & Vanilla NoRules \\
           & PaLM 2 & Metrics Reference \\
           & & Reference \\
           & & Swap \\
           & & Swap CoT \\
           & & Vanilla \\
           & & Vanilla NoRules \\
    \midrule
    LLMEval$^2$ & ChatGPT & Metrics Reference \\
             & & Vanilla NoRules \\
             & GPT-4 & Metrics Reference \\
             & & Vanilla NoRules \\
             & Llama 2 & Metrics Reference \\
             & & Vanilla NoRules \\
             & PaLM 2 & Metrics Reference \\
             & & Vanilla NoRules \\
    \midrule
    MT-Bench & ChatGPT & Metrics Reference \\
             & & Vanilla NoRules \\
             & GPT-4 & Metrics Reference \\
             & & Vanilla NoRules \\
             & Llama 2 & Metrics Reference \\
             & & Vanilla NoRules \\
             & PaLM 2 & Metrics Reference \\
             & & Vanilla NoRules \\
    \bottomrule
    \end{tabular}
    \caption{LLM evaluator modes used for the instruction following datasets in our experiments.}
    \label{tab:llm-eval-accuracy}
\end{table*}

\end{document}